\documentclass[conference]{ieeeconf}
\IEEEoverridecommandlockouts      % This command is only needed if
\usepackage[table, x11names]{xcolor}
\usepackage[colorlinks=true,linkcolor=black,citecolor=black,urlcolor=blue]{hyperref}
\usepackage{times} % assumes new font selection scheme installed
\usepackage{amsmath} % assumes amsmath package installed
\usepackage{amssymb}  % assumes amsmath package installed
\usepackage{graphicx}
\usepackage{algorithm}
\usepackage{mathtools}
\usepackage{algorithm}
\usepackage{algorithmicx}
\usepackage{float}
\usepackage{booktabs}
\usepackage[noend]{algpseudocode}
\usepackage[font=footnotesize]{caption}
\usepackage[font=footnotesize]{subcaption}
\usepackage[outdir=./]{epstopdf}

\definecolor{brewerGreen0}{HTML}{E5F5F9}
\definecolor{brewerGreen1}{HTML}{99D8C9}
\definecolor{brewerGreen2}{HTML}{2CA25F}

\definecolor{brewerCyan0}{HTML}{ECE2F0}
\definecolor{brewerCyan1}{HTML}{A6BDDB}
\definecolor{brewerCyan2}{HTML}{1C9099}

\definecolor{brewerGrey0}{HTML}{F0F0F0}
\definecolor{brewerGrey1}{HTML}{BDBDBD}
\definecolor{brewerGrey2}{HTML}{636363}

\definecolor{revisionColor}{HTML}{0238A8} %ia blue
\definecolor{lastRevisionColor}{HTML}{CC4C02} %ia orange

\usepackage{cite}

\usepackage{balance}
\usepackage{multirow} %ia required for information table

\usepackage{glossaries-extra}
\setabbreviationstyle[acronym]{long-short}
\glssetcategoryattribute{acronym}{nohyperfirst}{true}

\makeglossaries

\newacronym{slam}{SLAM}{Simultaneous Localization and Mapping}
\newacronym{sfm}{SfM}{Structure from Motion}
\newacronym{pgo}{PGO}{Pose-Graph Optimization}
\newacronym{vpr}{VPR}{Visual Place Recognition}
\newacronym{sgd}{SGD}{Stochastic Gradient Descent}
\newacronym{ils}{ILS}{Iterative Least-Squares}
\newacronym{gn}{GN}{Gauss-Newton}
\newacronym{lm}{LM}{Levenberg-Marquardt}
\newacronym{pcg}{PCG}{Preconditioned Conjugate Gradient}
\newacronym{map}{MAP}{Maximum-A-Posteriori}
\newacronym{gf}{GF}{Gaussian Filters}
\newacronym{pf}{PF}{Particle Filters}
\newacronym{sdp}{SDP}{Semi-Definite Programming}
\newacronym{bst}{BST}{Binary Search Tree}
\newacronym{ndt}{NDT}{Normal Distributed Transform}

\def\secref#1{Sec.~\ref{#1}}
\def\figref#1{Fig.~\ref{#1}}
\def\tabref#1{Tab.~\ref{#1}}
\def\eqref#1{Eq.~(\ref{#1})}

\def\ie{{i.e.}}
\def\eg{{e.g.}}
\def\etal{\emph{et al.}}
\def\lidar{LiDAR}
\def\lidars{LiDARs}

\usepackage{amsopn}

\newcommand{\bd}{\mathbf{d}}
\newcommand{\be}{\mathbf{e}}

\newcommand{\bp}{\mathbf{p}}
\newcommand{\bk}{\mathbf{k}}

\newcommand{\defeq}{=}

\DeclareMathOperator*{\argmin}{argmin}
\DeclareMathOperator*{\atantwo}{atan2}
\DeclareMathOperator*{\asin}{asin}

\def\g2o{$g^2o$}
\def\t2v{\mathrm{t2v}}
\def\v2t{\mathrm{v2t}}
\def\ev2t{\mathrm{ev2t}}

\newcounter{todonum}

\usepackage{lipsum}

\title{\LARGE \bf Visual Place Recognition using LiDAR Intensity Information}

\author{Luca Di Giammarino \and Irvin Aloise \and Cyrill Stachniss
  \and Giorgio Grisetti % <-this % stops a space
  \thanks{Luca Di Giammarino, Irvin Aloise, and Giorgio Grisetti are
    with the Department of Computer, Control, and Management
    Engineering "Antonio Ruberti", Sapienza University of Rome, Italy,
    Email:\,\,{\tt\footnotesize{\{digiammarino, ialoise,
        grisetti\}@diag.uniroma1.it}}.%
  }%
  \thanks{Cyrill Stachniss is with the University of Bonn,
  Germany, Email:\,\,{\tt\footnotesize{cyrill.stachniss@igg.uni-bonn.de}}
  }%
  \thanks{This work has partially been funded by the Deutsche
    Forschungsgemeinschaft (DFG, German Research Foundation) under
    Germany's Excellence Strategy, EXC-2070 -- 390732324 -- PhenoRob and
    from the European Union’s Horizon 2020 research and innovation 
    programme under grant agreement No 101017008 (Harmony).}%
}

\begin{document}
\maketitle
\thispagestyle{empty}
\pagestyle{empty}

%%%%%%%%%%%%%%%%%%%%%%%%%%%%%%%%%%%%%%%%%%%%%%%%%%%%%%%%%%%%%%%%%%%%%%%%%%%%%%%%
\begin{abstract}
Robots and autonomous systems need to know where they are within a map to navigate effectively. Thus, simultaneous localization and mapping or SLAM is a common building block of robot navigation systems. When building a map via a SLAM system, robots need to re-recognize places to find loop closure and reduce the odometry drift. Image-based place recognition received a lot of attention in computer vision, and in this work, we investigate how such approaches can be used for 3D \lidar{} data.  Recent \lidar{} sensors produce high-resolution 3D scans in combination with comparably stable intensity measurements. Through a cylindrical projection, we can turn this information into a panoramic image. As a result, we can apply techniques from visual place recognition to \lidar{} intensity data. The question of how well this approach works in practice has not been answered so far. This paper provides an analysis of how such visual techniques can be with \lidar{} data, and we provide an evaluation on different datasets. Our results suggest that this form of place recognition is possible and an effective means for determining loop closures.
\end{abstract}
%%%%%%%%%%%%%%%%%%%%%%%%%%%%%%%%%%%%%%%%%%%%%%%%%%%%%%%%%%%%%%%%%%%%%%%%%%%%%%%%
\section{Introduction} \label{sec:intro}

Robots need to perceive their surroundings to navigate safely and act effectively. \lidar{} sensors are a common sensor platform in robotics for several decades. Pushed by the increased safety required by the autonomous driving industry, 3D-\lidar{} technology rapidly evolved in recent years.  This resulted in having 3D instead of 2D sensing, fast and high-resolution point cloud acquisition, and intensity information for every 3D point -- all at a rather low cost. 

\begin{figure}[t]
	\centering
	\includegraphics[width=0.99\linewidth]{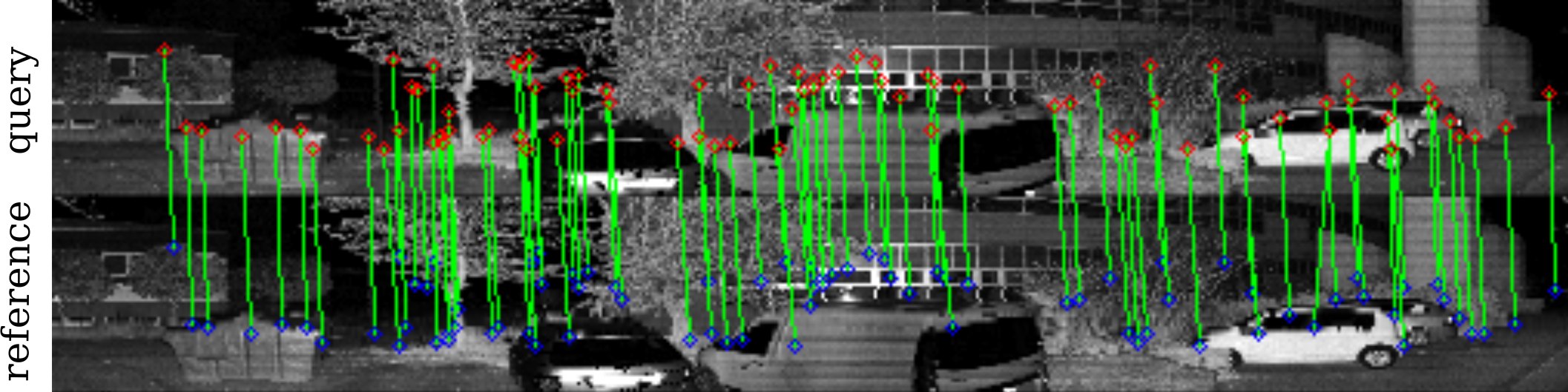}\\[1mm]
  \includegraphics[width=0.99\linewidth]{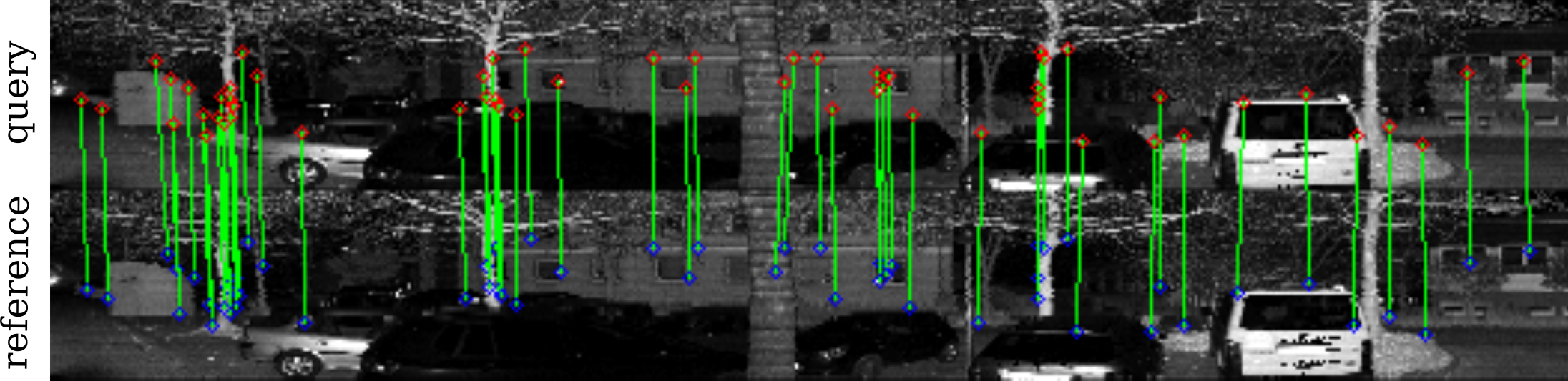}\\[1mm]
%	\noindent\rule{0.95\linewidth}{0.1pt}\\[2mm]
%	\vspace{10pt}
%	\medskip
	\hrulefill\par
	\vspace{8pt}
    \includegraphics[width=\linewidth]{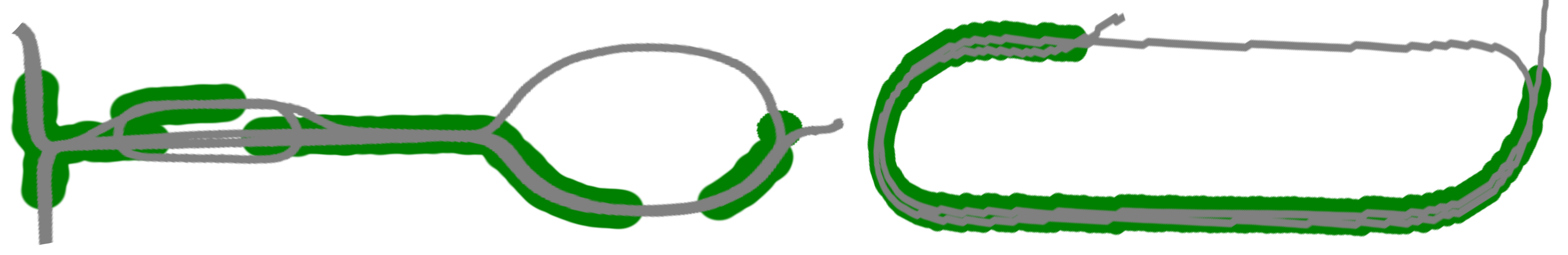}
    \caption{Today's 3D-\lidars{} measure both, the range and the intensity of the light reflected by the surrounding obstacles. Top: Example pairs of a query and a reference intensity image from our self-recorded dataset (\textit{IPB Car}). Note that due to its high horizontal resolution (original size: $64 \times 1024$), the intensity image  has been divided in two parts, one above the other. The green line illustrate descriptor matches provided by HBST~\cite{schlegel2018hbst} on intenisty data. Bottom: \textit{The Newer College} \cite{ramezani2020newer} (left) and \textit{IPB car} (right) datasets used for evaluation, valid loop closures using HBST highlighted in green.}
	\label{fig:motivation}
\end{figure}

%\todo{Add a seperating line between the matching image from the two datsets.}

Vehicles use \lidars{} to track their ego-motion as well as their surroundings
and to build point cloud maps of the scene. Most vehicles focus on the 3D information and 
rely on the well-known 
graph-based \gls{slam} paradigm to build maps. In this approach, the map of the environment is implicitly represented by the vehicle's trajectory, with point clouds or local maps attached to trajectory nodes.  
A graph-based SLAM system works by constructing a SLAM graph where each node represents a robot position, while edges encode a relative
displacement between nodes. These local transformations are inferred by comparing and matching nearby sensor readings.  Edges between subsequent robot positions can be straightforwardly estimated by registering point clouds incrementally~\cite{mendes2016icp, zaganidis2017semantic}. 
Relocalization events or so-called loop closures, occurring when the robot reenters in a known location after a long travel. The events should
trigger the creation of loop-closing edges, which are crucial to eliminate odometry drift and compute a globally consistent map. Finding such loop closures, however, can be challenging, especially in repetitive environments.

Existing 3D-\lidar{} \gls{slam} systems deliver accurate maps and yield
real-time performance on standard computers.  Still, a non-negligible number of approaches does not detect loops~\cite{zhang2014loam,zhang2015visual, deschaud2018imls, della2018general}, or rely on rather costly operations to check for candidate loop closures~\cite{behley2018efficient} -- \eg,~ICP in combination with outlier rejection mechanisms. 
This is because detecting loop-closures efficiently using only \lidar{} data is
still a challenge. Recently, learning-based techniques~\cite{chen2020rss} became popular as well. In the context of camera images, however, this task is framed as  \gls{vpr} and effective solutions 
exist~\cite{lowry2015visual,vysotska2019ral,naseer2018tro}.
In this paper, we investigate how such visual techniques can be applied to \lidar{} scans, especially the reflected intensity data, in an easy manner and how effective such an approach is. An example of the application of VPR approaches to \lidar{} intensity cues is depicted in \figref{fig:motivation}.

\clearpage

The main contribution of this paper is an analysis that evaluates how existing visual place recognition techniques perform when applied to the intensity cue of a 3D \lidar{} scanner. Thus, this paper is an experimental analysis and does not propose a new method to loop closing in general.  We tested several variants of \gls{vpr} pipelines in this context on multiple robotic datasets using 3D \lidars. Our experiments show that the straightforward adaptation of existing \gls{vpr} techniques  can produce reliable loop closures, enabling laser-only \lidar{} \gls{slam} at large scales.

%%%%%%%%%%%%%%%%%%%%%%%%%%%%%%%%%%%%%%%%%%%%%%%%%%%%%%%%%%%%%%%%%%%%%%%%%%%%%%%%
\section{Related Work}\label{sec:related}

The early loop-closures detection systems for 3D scans extracted features from the raw data.  Several
feature extractors have been proposed, each capturing some
traits of a local neighborhood of the scene.  Early studies in this
direction were made by Johnson~\cite{johnson1997spin} and later by
Huber~\cite{huber2002automatic}. The former extracted some local 3D
features from local point cloud patches, describing the local surface
around points with orientation. The latter built on top of Johnson's
Spin Images a methodology to perform global registration exploiting
these features. In this sense, each \textit{query} frame is compared
with a database, and if the surfaces of the local descriptors are
``similar'' between query and reference, then a potential loop-closure
is detected. Steder~\etal~\cite{steder2009robust} investigated novel
point features that are extracted directly from range images, and
later, they applied them in the context of loop-closures
detection~\cite{steder2010robust}. Finally, Steder~\etal~proposed to
use more robust NARF features~\cite{steder2011point} together with
Bag-of-Words-based search to increase the efficiency and the accuracy
of the detection~\cite{steder2011place}.  Orthogonally,
Magnusson~\etal~investigated the use of~\gls{ndt} as features to match
3D scans~\cite{magnusson2009automatic}. This approach has been
originally developed to perform registration between scans; still, the
authors demonstrated that \gls{ndt}-based features capture enough
structure to be used in the context of place
recognition. R{\"o}hling~\etal~\cite{rohling2015fast} investigated the
use of histograms computed directly from the 3D point cloud to define
a measure of the \textit{similarity} of two scans.  Novel types of
descriptors have been investigated, exploiting additional data gathered
by the \lidar{} sensor -- \ie,~light remission of the
beams~\cite{cop2018delight, guo2019local}. However, despite being very
attractive, these descriptors are time-consuming to extract and match,
resulting in a slower system overall.

More recently, deep learning approaches are spreading thanks to the increased computing power of today's computers. Dub{\'e}~\etal~\cite{dube2017segmatch} proposed the detection and matching of segments to recognize whether we are 
observing an already visited place. 
Uy~\etal~\cite{angelina2018pointnetvlad} employed a CNN based on 
PointNet~\cite{qi2017pointnet} to compute NetVLAD holistic 
descriptors~\cite{arandjelovic2016netvlad} out of range images. 
Zaganidis~\etal~\cite{zaganidis2019semantically} used semantic information 
extracted from the point cloud~\cite{qi2017pointnet++} to enrich \gls{ndt} 
features, resulting in more accurate and robust place recognition. 
Chen~\etal~\cite{chen2020rss}, instead, developed and end-to-end solution to 
evaluate the overlap of two 3D scans together with a raw estimate of the yaw 
angle.
Still, all deep-learning-based approaches require a great amount of data to 
perform training (most of the times also labeled) and a lot of computing power 
to work properly.

A lot of visual place recognition systems exploit features such as
SURF~\cite{bay2008cviu} or SIFT~\cite{lowe2004ijcv} and several approaches
apply bag-of-words techniques, i.e., they perform matching based on the
appearance statistics of such features. To improve the robustness of
appearance-based place recognition, Stumm \emph{et al.}~\cite{stumm2015icra}
consider the constellations of visual words and keeping track of their
covisibility.  Another popular approach for visual place recognition proposed by Galvez-Lopez \emph{et al.}~\cite{galvez2012bags} proposes a bag of words approach using binary features for fast image retrieval.
Single image visual localization in real-world outdoor environments is still an active field of research, and one popular
approach used in robotics is FAB-MAP2~\cite{cummins2009rss}. 
For across season matching using SIFT and SURF, 
Valgren and Lilienthal~\cite{valgren2010jras} propose to  combine features and
geometric constraints to improve the matching. 

To deal with substantial variations in the visual input, it is
useful to exploit sequence information for the alignment,
compare~\cite{linegar2015icra,milford2013ijrr,milford2012icra,naseer2018tro,vysotska2016ral, vysotska2019ral}.
SeqSLAM~\cite{milford2012icra}  aims at matching image sequences under seasonal changes and computes a matching matrix that
stores the similarity between the images in a query sequence and a database. Milford \emph{et al.}~\cite{milford2013ijrr}   present a
comprehensive study about the SeqSLAM performance on low-resolution images.
Related to that, Naseer \emph{et al.}~\cite{naseer2018tro} focus on 
sequence matching using a network flow approach and Vysotska \emph{et
al.}~\cite{vysotska2016ral} extended this idea towards an online approach with
lazy data association and build up a data association graph online on-demand, also
allowing flexible trajectories in a follow-up work~\cite{vysotska2019ral}.

In this paper, we investigate how to perform fast and accurate 
place-recognition using additional channels available in modern 3D-\lidar{} 
sensors. Our approach applies well-known methodologies originally designed to 
work with camera images to 3D-LiDARs data, exploiting the increased 
the descriptiveness of such sensors. We perform multiple experiments with different combinations of features - image retrieval tools. Among the features, we picked computationally efficient binary ones like BRISK \cite{leutenegger2011brisk} and ORB \cite{rublee2011orb}. Instead, as floating point descriptors, we selected SURF~\cite{bay2008cviu} and Superpoint, a more recent neural extractor that shows impressive results compared to older geometrical features~\cite{detone2018superpoint}. Among image-retrieval tools, we use HBST \cite{schlegel2018hbst}, a tree-like structure that allows for descriptor search and insertion in logarithmic time by exploiting particular properties of binary feature descriptors, and DBoW2~\cite{galvez2012bags} that allows fast image retrieval based on the histogram of the distribution of words appearing in the image both for floating point and binary descriptors.

%%%%%%%%%%%%%%%%%%%%%%%%%%%%%%%%%%%%%%%%%%%%%%%%%%%%%%%%%%%%%%%%%%%%%%%%%%%%%%%%
\section{LiDAR Sensors in Robotics}\label{sec:modern-lidars}
\begin{figure}
	\centering
	\includegraphics[width=0.99\linewidth]{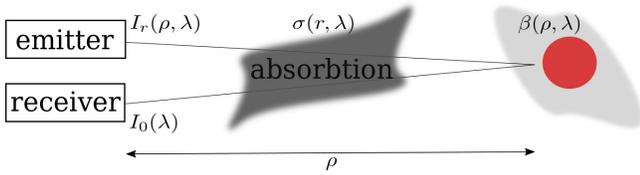}
	\caption{Illustration of a single beam emitting and receiving light pulse, as described in \eqref{eq:laser-model}.}
\label{fig:single-beam-shit}
\vspace{-0.30cm}
\end{figure}

A typical LiDAR sensor emits a beam of pulsed light waves towards the
measurement direction. The distance to the obstacle along the beam is
measured from the light pulse's round trip time.  At a low level, a
\lidar{} senses the perceived light intensity $I_r(\rho, \lambda)$ as a
function of the range of the reflection $\rho$ and the wavelength
$\lambda$. The sensed intensity depends on the emitted intensity
$I_0(\lambda)$ at the same wavelength, as follows:
\begin{equation}
 I_r(\rho, \lambda) = 
 I_0 \eta \frac{A}{4  \pi \rho ^2} \beta(\rho, \lambda)
 \exp\left(-2 \int_0^\rho \sigma(r, \lambda) \;dr\right).
 \label{eq:laser-model}
\end{equation}
Here, $A$ denotes the beam aperture measured as a solid angle, $\beta$
is the reflectance of the object, and $\sigma$ is the absorption of
the medium.  \figref{fig:single-beam-shit} illustrates this aspect.
The reflectivity $\beta$ is affected by the composition, roughness and
moisture content and incidence angle of the beam hitting the surface.

The resolution of the clock measuring the first return of the signal
bounds the range measurement's resolution. Additional
accuracy gains, however, can be obtained by determining the phase difference between the emitted and received signal. A single detection can also be rather noisy. Therefore, scanners targeting higher accuracies send multiple pulses. This, in turn, caps the frequency of range measurements. The measurement frequency~$f_m$ of a single sensor nowadays might reach 50\,kHz. By choosing the rotation frequency $f_r$ of the beam, the angular resolution is straightforwardly $2\pi \frac{f_m}{f_r}$.
Mechanical considerations limit the rotational speed of the sensor.
Whereas in the 2D case, the beam can be deflected by a
relatively small rotating mirror, the sensor head carries multiple
measuring units in the 3D case -- today up to 128.  

Scanners also measure the intensity of measurements as the amount of light reflected from the surface. This intensity information is normalized and discretized to an 8 or 16 bit value. The intensity depends on surface characteristics, and several other factors impact the measurement. All terms in \eqref{eq:laser-model} are continuous. Thus, we can expect that mild changes of the viewpoint yield mild variations of the intensity when measuring the same 3D point.

The majority of prior works on 3D-\lidars{} for SLAM and place recognition
focused on using the range measurements and ignored other potentially valuable information~\cite{zhang2015visual, droeschel2018efficient}. This may also be since the intensity information of older scanners used in robotics was not as great. Compared to their predecessors, however, recent
3D-\lidars{} exhibit an increased accuracy and vertical resolution.  When assembled in a panoramic image, the intensity recalls the
one obtained by using a grayscale camera. Undoubtedly, the quality of a \lidar~intensity image is still low compared to the one acquired by passive sensors such as cameras. However, in robotics and specifically in \gls{vpr} tasks, the intensity image generated from a 3D-\lidar~scan brings advantages such as invariance to external light conditions and shadows.

The popularity of autonomous driving and self-driving cars pushed the
improvement of 3D-\lidars{}. In this application domain, their main use is
to provide local 3D reconstruction and obstacle information.
Traditionally, global 3D reconstruction using \lidars{} presents a
significant challenge of loop-closing. This arises from the higher
sensor aliasing between range only 3D scans, compared to more
descriptive images.  The literature is rich in registration and mapping
algorithms for 3D-\lidars{}, whereas this community invested less effort
in tasks such as place recognition, which forms the basis for
effective loop closing.  In contrast, the computer vision community
invested substantial efforts in this place recognition task achieving impressive results~\cite{lowry2015visual}.  Therefore, the purpose of this paper is to analyze the behavior of common VPR approaches when used in combination with \lidar{} intensity information.

%%%%%%%%%%%%%%%%%%%%%%%%%%%%%%%%%%%%%%%%%%%%%%%%%%%%%%%%%%%%%%%%%%%%%%%%%%%%%%%%
\section{Visual Place Recognition Applied \\ to Cylindrical \lidar{} Intensity Images}\label{sec:our-approach}

Popular \gls{vpr} approaches often store a database of places in the form of a collection of image keypoints and descriptors (and potentially a coordinate in some world frame). The \textit{keypoints} are \textit{salient points} in the image, possibly corners and edges, while the \textit{descriptors} encode the \textit{appearances} around keypoints. 

In the process of finding similar places, two images are regarded as similar 
if a substantial part of their keypoints' descriptors are close to each other. 
Performing \gls{vpr} using this paradigm
requires first to convert a query image into a set of keypoints and
descriptors and second to efficiently find images with similar descriptors. Effective solutions are
available to quickly find the potential matches in the database, see~\cite{lowry2015visual}. Among all, we focus specifically on HBST~\cite{schlegel2018hbst} and DBoW2~\cite{galvez2012bags} as two promitent approaches.

An intensity image constructed from a laser scan has a number of rows
equal to the number of vertical beams and a number of columns equal to
the number of scanning steps along the azimuth.  Unfortunately, most
public datasets provide the scans as annotated point clouds, and
recovering the beam measurements needed for image formation requires
a cylindrical projection. Due to vehicle motion, round-offs, or unknown parameters,
this projection will likely result in missing data in some parts of the image.
These phenomena may hinder the straightforward feature extraction process.

In the following, we will first discuss how we handle image formation from
a laser scan, and then we review the structure of a straightforward pipeline for \gls{vpr}.

\subsection{Image Formation}
\label{sec:image-formation}

When data of the raw beam measurements are not available given the scanner setup or given the dataset, we can compute a cylindrical image from the 
3D point cloud $\mathcal{P}_\mathrm{lid}$ by spherical projection:
$\Pi : \mathbb{R}^3 \mapsto \mathbb{R}^2$.
This is done by converting each Cartesian point
$\bp = \left(x, y, z\right)^\top \in \mathcal{P}_\mathrm{lid}$
as a spherical one $\bar\bp = 
\left(\rho, \theta, \phi\right)^\top$, with:
\begin{align*}
  \rho &\defeq \sqrt{x^2 + y^2 + z^2} \\
  \theta &\defeq \atantwo(y,x) \in \left[-\pi / 2, \pi / 2\right] \\
  \phi &\defeq \asin(z / \rho) \in \left[-\pi, \pi\right].
\end{align*}

Assuming that the beams are \emph{uniformly spaced} over $f$, 
we can compute $\left(u, v\right)^\top$ as follows:
\begin{equation}
  \begin{pmatrix}
    u \\ v
  \end{pmatrix} \defeq 
  \begin{pmatrix}
    \frac{W}{2} \left( 1 + \frac{\theta}{\pi} \right) \\
    H \frac{f_\mathrm{up}-\phi}{f_\mathrm{up}-f_\mathrm{low}} 
  \end{pmatrix},
  \label{eq:spherical-projection}
\end{equation}

\noindent
where $W$ and $H$ are respectively the width and height of the image and $f = f_\mathrm{up} + f_\mathrm{down}$ is the vertical FoV of the sensor.
Should multiple points fall in the same image pixel, only the value having the smallest range is retained. In each pixel of the created image, we store the intensity value and not the range.

\begin{figure}
	\centering
	\includegraphics[width=0.99\linewidth]{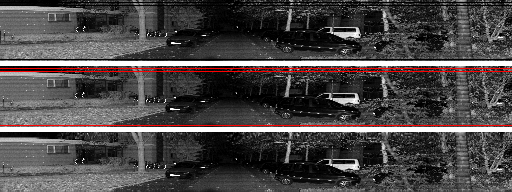}
	\caption{Empty lines removal. From top to bottom, original image after projection from 3D point cloud as explained in \secref{sec:image-formation}, detection of empty rows highlighted in red, result after image manipulation. Each image shown has been cropped to half of their horizontal size for better viewing.}
	\label{fig:line-removal}
	\vspace{-0.4cm}
\end{figure}

The uneven distribution of the vertical beams
may lead to empty gaps in the resulting image, usually whole
horizontal rows. This problem can be solved a posteriori by either
scaling down the vertical resolution or by performing interpolation.  In the experiments on this paper, we adopted the second
method.  To remove the empty rows from a panoramic image, we first detect
them using a binary threshold and a horizontal kernel as wide as
the image.  We compute the interpolated value for each pixel in the empty rows through bilinear interpolation based on the
upper and lower valid rows' values.
\figref{fig:line-removal} shows the result of this procedure.
\subsection{Feature Extraction}\label{sec:feature-extraction}
As stated at the beginning of this section, the feature extraction process aims at compressing an image in a set of interest points
or \textit{keypoints}.  A \textit{descriptor} vector captures the appearance of the image in the neighborhood of the keypoint.
The detector outputs a set of keypoint $\{\bk_i = \left(u_i, v_i\right)^\top \}$,
in image coordinates.
A key quality of a keypoint detector is its ability to identify points
that are ``salient'' or ``locally distinct''. In other words, a 
good detector will
identify the projection of the same point in the world upon small changes
in the viewpoint. Typical approaches consider the image gradient at different scales to compute keypoints. Thus, to successfully operate, a
detector requires the gradients in the image to capture the local
intensity difference at nearby regions of the world. Accordingly,
these approaches do not work when the vertical resolution is too low,
since in this case, changes in the gradient are dominated by sampling
effects.
\begin{table}[!ht]
	\begin{center}
		\begin{tabular}{c c c} 
			\toprule
			FAST & threshold & 40 \\
			\midrule
			ORB & nFeatures & 300 \\
			&  scaleFactor & 1.2 \\
			&  scaleFactor & 8 \\
			&  nLevels & 8 \\
			&  edgeThreshold & 15 \\
			\midrule
			BRISK & threshold & 30 \\   
			& octaves & 3 \\
			& patternScale & 1 \\
			\midrule
			SURF & hessianThreshold & 400 \\    
			& nOctaves & 4 \\      
			& nOctaveLayers & 3 \\
			\midrule
			Superpoint & minProbability & 0.05 \\
			& nFeatures & 300 \\
			\bottomrule
		\end{tabular}
		\caption{\label{tab:configurations} Configuration of keypoints detector and descriptors extractors used.}
	\end{center}
\vspace{-0.7cm}
\end{table}
Similarly, typical feature detectors operate on a small image patch
from which they compute some quantity that is as invariant as possible
to mild warpings of the patch itself. This ensures that regions of the
image that look alike will result in similar descriptors.  For each
keypoint $\bk_i$, the extractor computes a descriptor vector
$\bd(\bk_i)$. This vector consists of either floating point or binary values.

We directly employed well-known combinations of feature detectors and
extractors\cite{leutenegger2011brisk, rublee2011orb, bay2006surf}
whose C++ implementation is publicly
available~\cite{bradski2000opencv}. We also tested a more recent
neural feature extractor by Detone et al.~\cite{detone2018superpoint}.
\begin{figure}[!ht]
	\centering
	\includegraphics[width=\linewidth]{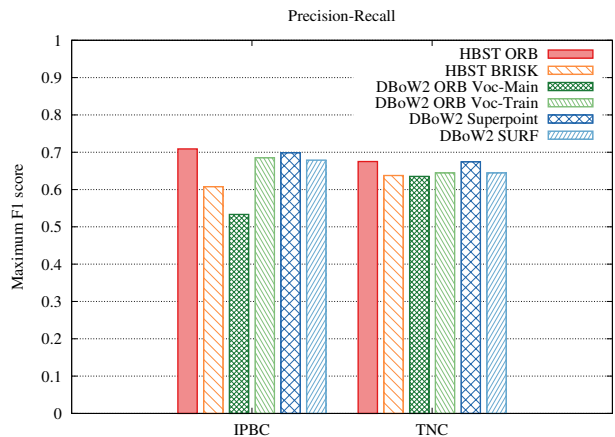}
	\caption{Max $F_1$ Score reached in full evaluation mode. Left our recorded dataset \textit{IPB Car} (self-recorded) and right \textit{The Newer College} \cite{ramezani2020newer}.}
	\label{fig:f1-score}
	\vspace{-0.3cm}
\end{figure}
\begin{table*}[!ht]
	\begin{center}
		\begin{tabular}{c c c c c c c} 
			\toprule
			Dataset & Description & \lidar{} Model & Vertical FoV [$\deg$] & Ground Truth\\
			\midrule
			\textit{The Newer College} (seq 00) \cite{ramezani2020newer} & outdoor dynamic, campus-park & OS1-64 & 45 & Ext. Localization System\\
			\midrule
			\textit{IPB Car} (self-recorded) & outdoor dynamic, urban & OS1-64 & 45 & RTK-GPS\\
			\midrule
			\textit{Ford Campus} (seq 00) \cite{pandey2011ford} & outdoor dynamic, urban&  HDL 64-E & 26.9 & RTK-GPS\\
			\midrule
			\textit{KITTI} (seq 00) \cite{geiger2013vision} & outdoor dynamic, urban & HDL 64-E & 26.9 & RTK-GPS\\
			\bottomrule
		\end{tabular}
		\caption{\label{tab:datasets} Datasets we used for evaluation.}
	\end{center}
\vspace{-0.70cm}
\end{table*}

\subsection{Feature-Based \gls{vpr}}\label{sec:vpr}
Two images of the same scene acquired with similar viewpoints
will have a high number of descriptors that have a small distance.
To this extent, we should define a 
suitable metric $\be_d$ for this comparison. 
For floating point descriptors, $\be_d$ 
a standard choice is the Euclidean distance in $\mathbb{R}^n$ (other metrics such as the $\cos$-similarity could be employed instead). For binary ones, the Hamming distance is commonly employed.

Relying on the metric $\be_d$ and the invariant properties of the
descriptors, we can find corresponding points between two images
$\mathcal{I}_q$ and $\mathcal{I}_r$ by, finding for each keypoint
$\bk_q \in \mathcal{I}_q$ the keypoint $\bk_r \in \mathcal{I}_r$ that
has minimal distance in the descriptor space:
\begin{equation}
  \bk_r^\star = \argmin_{\bk_r}\left(\be_d(\mathbf{d}(\bk_q), 
  \mathbf{d}(\bk_r))\right) \; : \bk_q \in \mathcal{I}_q \quad \bk_r \in 
  \mathcal{I}_r.
  \label{eq:descriptor-matching}
\end{equation}

A straightforward way to solve \eqref{eq:descriptor-matching} is by
\textit{exhaustive search}. This process is complete since it returns all neighbors according to the distance metric.  However, it quickly becomes
prohibitive as the size of the database increases, thus preventing online operations.
Efficient approaches that perform an approximate search
are available. These methods usually organize the features in the database
in a search structure.
Common choices are search trees such as KD-trees or binary trees.
The splitting criterion and the parameters of the tree control
the completeness of the search.

Alternative methods preprocess the features in the image by describing
each image as a histogram of ``words''. The words are computed by
determining a priori a ``dictionary'' from a training image set.
The elements of the dictionary are the clusters of features in the training set.
Each feature in an image will contribute to its histogram based on the ``word''
in the dictionary closest to the feature.
As a representative for tree-based approaches, we use HBST~\cite{schlegel2018hbst},
while for BoW, we used DBoW2~\cite{galvez2012bags}. In the next section, we will discuss in more detail the experimental configuration.

\begin{figure*}[t]
  \centering
  \includegraphics[width=0.96\linewidth]{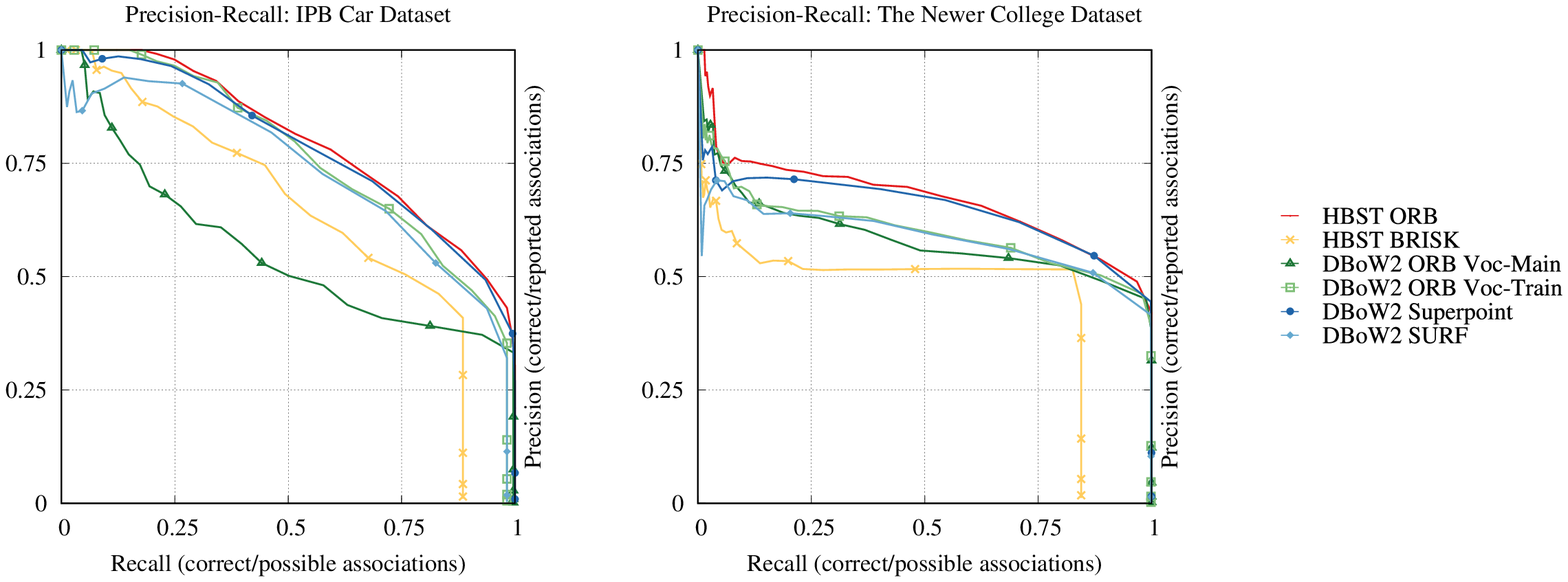}
  \caption{Precision-Recall curves of the closures computed both with different combinations of feature extractors - image matchers on the \lidar~ intensity image. Greater accuracy is reported in general by ORB-HBST, ORB-DBOW2 and Superpoint-DBOW2.}
  \label{fig:precision-recall}
%\end{figure*}
\vspace{5mm}
%\begin{figure*}[t]
  \centering
  \includegraphics[width=0.96\linewidth]{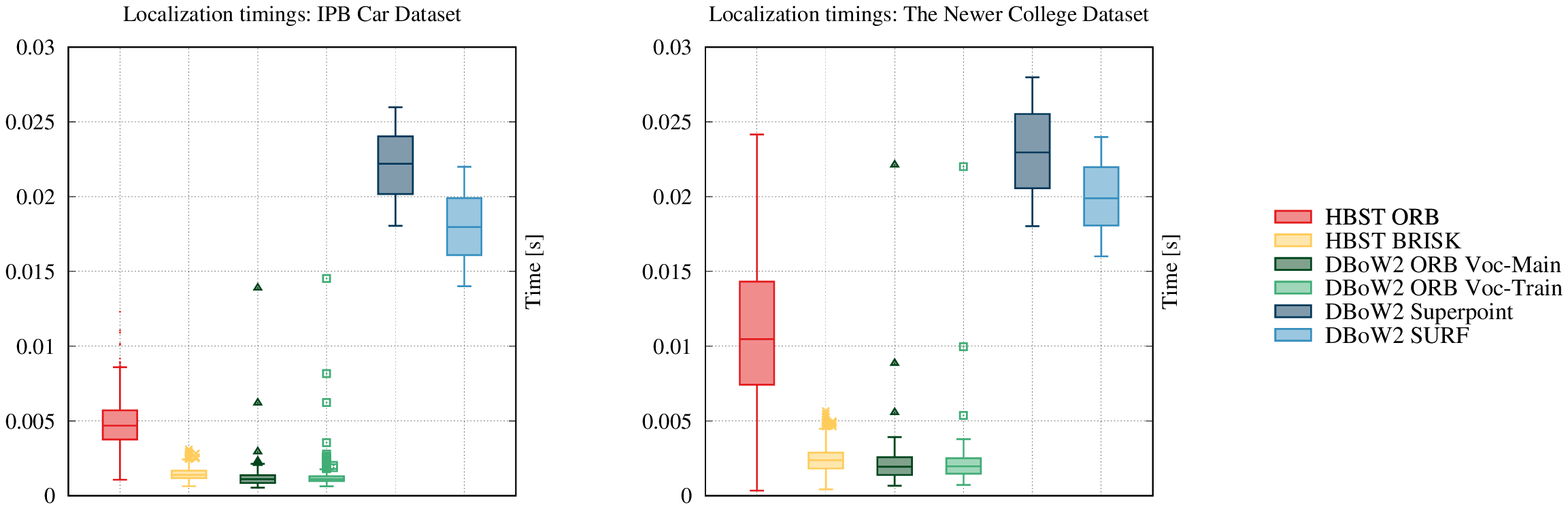}
  \caption{Localization timings of different combinations of feature extractors - image matchers. As expected binary descriptors take lower time to extract and match compared to floating points one.}
  \label{fig:loc-timings}
\end{figure*}

\section{Experimental Evaluation}\label{sec:experiments}

This evaluation analyzes our implemented combination of feature extractors and \gls{vpr} pipelines on intensity images generated from \lidar{} scanner point clouds and intensity data. In more detail, we evaluate the following combinations:

\begin{itemize}
	\item FAST - ORB - HBST
	\item FAST - BRISK - HBST
	\item FAST - ORB - DBoW2
	\item Superpoint - DBoW2
	\item FAST - SURF - DBoW2
\end{itemize}  
	
We test these configurations on four datasets, three publicly available, and one self-recorded dataset using an automated car, see \tabref{tab:datasets} for dataset details. All datasets provide some cue of the robot position independent from the \lidar{} sensor, which we use to construct ground truth place information. This is usually done with an RTK-GPS or an external reference system.
The \textit{ground truth} is a set of matching pairs of scans acquired at nearby locations.
A pair is a match if the scans' recording locations are close according to the external system, and an ICP registration succeeds. We processed the datasets sequentially by adding the query image $\mathcal{I}_q$ to the database at each step. We obtain a set (potentially empty) of images similar to $\mathcal{I}_q$ at each query, and we verify these matches against the ground truth.  From the returned images, we disregard those added within the most recent 120 steps. We do this to not positively bias the evaluation.

For each dataset, we tested the combination of feature extractor and image retrieval systems mentioned before.
To quantify the performances of one run, we 
evaluated common statistical quantities - \ie,~\emph{Precision} and 
\emph{Recall}. To this end, we introduce the terms \emph{true positive} to 
indicate a loop-closure that is present in the ground truth database and 
\emph{false positive} to indicate a wrong loop-closure. Analogously, a 
\emph{false negative} represents a loop-closure that is 
present in the ground truth database but has not been reported by the method in 
analysis; a \emph{true negative} represents its contrary.
Hence, we can define Precision and Recall using the number of true positives 
$T_p$, false positives $F_p$, true negatives $T_n$ and false negatives $F_n$ as 
follows:

\begin{equation}
  P = \frac{T_p}{T_p + F_p} \qquad R = \frac{T_p}{T_p + F_n}.
  \label{eq:precision-recall}
\end{equation}

We use FAST as the keypoint detector apart from Superpoint that outputs pairs of keypoints and descriptors directly for all experiments. For each dataset, we extracted the following descriptors: ORB,
BRISK as binary and Superpoint and SURF as floating point.  As
retrieval methods, we use HBST with parameters $\delta_\mathrm{max} = 0.1$ and $N_\mathrm{max} = 50$ for the
binary features. We use DBoW2 for all features but BRISK, which is not supported in by default package, and we extend it to operate with Superpoint. Configuration of each feature extractor reflects \tabref{tab:configurations}.

Since BoW approaches require a dictionary that depends on the sensor characteristics, we train such dictionaries using a portion of the datasets not used for the evaluation. In the plots, these curves are labeled with Voc-Train. More in detail, we train the vocabulary by using 3000 intensity images, a branching factor of 10, and a depth level of 5, using the classic euclidean distance to measure descriptor similarity over floating points and the Hamming distance for the binary ones.
For comparison, we also report the results obtained with the image-based
dictionaries packaged in the software release of DBoW2 (Voc-Main).

We conduct several experiments, varying the type of descriptor
and \gls{vpr} matcher.  We report the precision-recall curves (\figref{fig:precision-recall}), maximum harmonic mean (\tabref{tab:f1-score}, \figref{fig:f1-score}), localization timings (\figref{fig:loc-timings}) and valid loop closures drawn on the trajectories (\figref{fig:f1-score}) of the most significant experiments.  
\\
 
Overall, existing \gls{vpr} approaches shows usable results at
negligible computation over the two most recent datasets (\textit{IPB Car} and \textit{The Newer College} \cite{ramezani2020newer}) (\figref{fig:precision-recall} - \figref{fig:loc-timings}). Results obtained in \textit{KITTI} \cite{geiger2013vision} and \textit{Ford Campus} \cite{pandey2011ford} are not comparable with the other two modern datasets. The vertical resolution and the intensity quality were to low in \textit{KITTI} and \textit{Ford Campus} to detect stable features, and we were not able to produce acceptable results (\tabref{tab:f1-score}).
\begin{table}[!ht]
	\begin{center}
		\begin{tabular}{c c c c} 
			\toprule
			The Newer College \cite{ramezani2020newer} & IPB Car & Ford Campus \cite{pandey2011ford} & KITTI \cite{geiger2013vision} \\
			\midrule
			0.6751 & 0.7088 & 0.115 & 0.097 \\ 
			\bottomrule
		\end{tabular}
		\caption{\label{tab:f1-score} Max $F_1$ score reached in full validation over the four datasets with combination of HBST \cite{schlegel2018hbst} and ORB \cite{rublee2011orb}.}
	\end{center}
\vspace{-0.3cm}
\end{table}

We obtain the best results by combining ORB with HBST and DBoW2. The combination Superpoint-DBoW2 shows a comparable performance.  However, floating point descriptor comes with a higher computational cost, see \figref{fig:loc-timings}. The
accuracy on \textit{The Newer College} dataset \cite{ramezani2020newer} is inferior to the one
obtained on our self-recorded dataset called \textit{IPB Car}. This is due to the small changes on the roll axis since this data has been recorded walking in the campus, which, in turn, translates into a higher viewpoint variation within the same dataset.  

For the experimental campaign, we used a PC running Ubuntu
20.04, equipped with an Intel i7-10750H CPU@2.60GHz and 16GB of RAM. We run neural network-based feature detection on an NVIDIA GeForce GTX 1650Ti.
%%%%%%%%%%%%%%%%%%%%%%%%%%%%%%%%%%%%%%%%%%%%%%%%%%%%%%%%%%%%%%%%%%%%%%%%%%%%%%%%
\section{Conclusion} \label{sec:conclusion}

In this work, we provide an analysis of the performance of visual place recognition techniques for loop closing, applied to the intensity information of a 3D \lidar{} scanner. We evaluated gold standard visual place recognition approaches on four different datasets. Except for one outdated \lidar{}, which does not provide stable intensity measurements,  this transfer was proved to be successful. On modern sensors with a high vertical resolution, we obtained encouraging results. Despite not proposing a new approach in this work, we believe that existing \lidar{}-based mapping systems can easily benefit from our findings. We furthermore expect that one can improve the performance further by designing or learning descriptors that are specifically optimized for  intensity cues of \lidar{} scanners.

%%%%%%%%%%%%%%%%%%%%%%%%%%%%%%%%%%%%%%%%%%%%%%%%%%%%%%%%%%%%%%%%%%%%%%%%%%%%%%%%

\bibliographystyle{plain}
\bibliography{robots}
\balance

\end{document}